# Updating with belief functions, ordinal conditional functions and possibility measures

Didier DUBOIS     Henri PRADE

*Institut de Recherche en Informatique de Toulouse*
*Université Paul Sabatier, 118 route de Narbonne*
*31062 TOULOUSE Cedex (FRANCE)*

**Abstract** : This paper discusses how a measure of uncertainty representing a state of knowledge can be updated when a new information, which may be pervaded with uncertainty, becomes available. This problem is considered in various framework, namely : Shafer's evidence theory, Zadeh's possibility theory, Spohn's updating approach. In the two first cases, analogues of Jeffrey's rule of conditioning are introduced and discussed. The relations between Spohn's model and possibility theory are emphasized and Spohn's updating rule is contrasted with the Jeffrey-like rule of conditioning in possibility theory.

## 1 - Belief updating in an uncertain environment

One of the strong assets of probability theory for reasoning with uncertain information is the existence of Bayes rule of conditioning that serves as a basis for an efficient theory of belief updating. This feature is apparently missing in alternative theories of uncertainty such as belief functions and possibility theory, despite the existence of conditioning notions that are more and more discussed currently. Especially whether Dempster rule and other symmetric combination rules can serve as a substitute to Bayesian inference is debatable. In this paper we make a step toward addressing these problems by proposing alternative updating rules that extend the notion of conditioning while preserving the intrinsic dissymmetry of the process of updating knowledge bases.

Here knowledge bases are considered from a *semantic* point of view. This means that their contents are supposed to be represented by a unique weight distribution on a suitable universe, in a given uncertainty modeling framework. In the following, Shafer (1976)'s basic probability assignments, Zadeh (1978)'s possibility distributions, Spohn (1988, 1989)'s ordinal conditional functions are the various candidates considered for the representation of the contents of a knowledge base. Thus when discussing updating issues, we do not take into account the existence of *syntactically* different knowledge bases having the same semantic description of their contents. The syntactic view would enable us to take care of the origin of each piece of information in a revision process. For instance let us consider the simple case of two knowledge bases (not pervaded with uncertainty) {A, A→B} and {A,B} which are semantically equivalent since they have the same set of consequences, namely the consequences of A ∧ B ; then we may be led to revising them in different ways when receiving the new information ¬A. Although more simple, the semantic view raises problems which are the topic of this paper. Indeed, most of the available combination rules (e.g. Dempster's rule, Zadeh's min-combination) are symmetrical ; thus if we use them for combining the distribution expressing the contents of the whole knowledge base with the one representing the new information, we consider the old and the new information at the same level, which is debatable.

The next two sections discuss conditionalization operations for updating purposes in Shafer's evidence theory and Zadeh's possibility theory respectively. These operations are not symmetrical and can be regarded as analogues of Jeffrey (1965)'s rule in probability theory. Then it is shown that Spohn's ordinal conditional functions are equivalent to possibility distributions and then conditionalization operations introduced by Spohn can be compared to the ones studied in the two other frameworks (since mathematically speaking, possibility measures can be regarded as particular cases of Shafer's plausibility functions).

## 2 - Updating in Shafer's evidence theory

In probability theory, conditioning is defined by Bayes' formula

This work was partially supported by the European ESPRIT Basic Research Action number 3085, entitled "Defeasible Reasoning and Uncertainty Management Systems" (DRUMS).

$$P(B \mid A) = \frac{P(A \cap B)}{P(A)} \qquad (1)$$

where $P(\cdot)$ denotes the prior probability, A is observed (with complete certainty) and $P(\cdot \mid A)$ denotes the a posteriori probability measure, taking A for granted. A and B are supposed to be subsets of a referential set $\Omega$.

Jeffrey (1965)'s rule extends Bayes' conditioning to the case where the observation is pervaded with uncertainty. Let $\alpha$ be the (probabilistic) certainty with which A is observed, and thus $1 - \alpha$ corresponds to the certainty that $\bar{A}$ is actually observed. Then the updated probability measure P' is defined by

$$P'(B) = \alpha \cdot P(B \mid A) + (1 - \alpha) \cdot P(B \mid \bar{A}) \qquad (2)$$

where $P(B \mid A)$ and $P(B \mid \bar{A})$ are given by (1). This expression is generalized to the case where the possible observations $A_1, ..., A_n$ make a partition, and where $\alpha_i$ is the certainty of having observed $A_i$ (with $\Sigma_{i=1,n} \alpha_i = 1$), by

$$P'(B) = \Sigma_{i=1,n} \alpha_i \cdot P(B \mid A_i) \qquad (3)$$

In his book Pearl (1988) tries to cast this rule within the classical Bayesian framework, noticing that $\alpha_i$ could be interpreted as a conditional probability $P(A_i \mid E)$ where E denotes the event producing an uncertain observation. Then P'(B) is of the form $P(B \mid E)$ provided that E and B are conditionally independent given $A_i$, for all i. See Shafer (1981) for a comparison with Dempster's rule of combination.

Nevertheless let us recall that the linear convex combination is the unique way of combining probability measures in an *eventwise* manner (the same combination law applies for each event) which leads to a probability measure as a result (Lehrer & Wagner, 1981 ; Berenstein et al., 1986). Thus the expression (3), whatever its other justifications, is not at all surprising.

Lastly note that if $\Omega = \{\omega_1, ..., \omega_m\}$ and $A_i = \{\omega_i\}$, $\forall i = 1,m$, $\alpha_i = P_2(\{\omega_i\})$ for a probability measure $P_2$, then Jeffrey's rule (3) comes down to a simple substitution of the prior probability P by the uncertain observation $P_2$, i.e. $P'(B) = P_2(B)$, $\forall B$, since $P(B \mid \{\omega_i\}) = 1$ if $\omega_i \in B$, and 0 otherwise.

Let us now consider Shafer's evidence theory. In this framework the available knowledge is represented in terms of a basic probability assignment m, which is a set function from the set of subsets $2^\Omega$ of a so-called frame of discernment $\Omega$ to [0,1] with the constraints $m(\emptyset) = 0$ and $\Sigma_A m(A) = 1$. The subsets $A \subseteq \Omega$ such that $m(A) > 0$ are called focal elements. Note that there is no constraint on the structure of the set $\mathcal{F}$ of focal elements (here supposed finite and which does not make a partition in general). Let us emphasize that the pair $(\mathcal{F},m)$ can be viewed as a random set (Goodman & Nguyen, 1985 ; Dubois & Prade, 1986a). This means that each focal element $A_i$ represents the most accurate description of the reality with certainty $m(A_i)$. The subsets $A_i$ are the possible realizations of the observation pervaded with uncertainty. Due to the incompleteness of the available information, $A_i$ is not necessarily a singleton. A plausibility function Pl as well as a belief function Bel can be bijectively associated with m (Shafer, 1976) and are defined by

$$Pl(B) = \Sigma_{A: A \cap B \neq \emptyset} \, m(A) \qquad (4)$$

$$Bel(B) = 1 - Pl(\bar{B}) = \Sigma_{\emptyset \neq A \subseteq B} \, m(A) \qquad (5)$$

In terms of plausibility functions, Dempster rule of conditioning is expressed by

$$Pl(B \mid A) = \frac{Pl(A \cap B)}{Pl(A)} \; ; \; Bel(B \mid A) = 1 - Pl(\bar{B} \mid A) \quad (6)$$

This rule of conditioning can be justified on the basis of an axiom that defines a conditional function associated to any set-function f defined on $\Omega$ as follows (Cox, 1946) :

$$f(A \cap B) = f(A \mid B) * f(B) \qquad (7)$$

which expresses that the degree attached to $A \cap B$ is a function $*$ of the degree attached to B combined with the degree attached to A, given that B is taken for granted. It is well known that the Boolean structure of $2^\Omega$ forces $*$ to be a product up to an isomorphic transformation, when $*$ is strictly monotonic in both places (e.g. Cox, 1946 ; Aczel, 1966). Note that (7) justifies Dempster's conditioning rule as well as the geometric rule of conditioning (Suppes & Zanotti, 1977)

$$Bel_g(A \mid B) = \frac{Bel(A \cap B)}{Bel(B)}; \; Pl_g(A \mid B) = 1 - Bel_g(\bar{A} \mid B) \quad (8)$$

In terms of basic probability assignments, $Pl(\cdot \mid B)$ defined by (6) is obtained by transferring all masses m(A) over to $A \cap B$, followed by a normalization step, while $Bel_g(\cdot \mid B)$ is obtained by letting $m_g(A \mid B) = m(A)$ if $A \subseteq B$ and 0 otherwise, followed by normalization, i.e. a more drastic way of conditioning (see Dubois & Prade, 1986b). Dempster's rule of conditioning looks more attractive from the point of view of updating since $Pl(A \mid B)$ is undefined only if $Pl(B) = 0$ (i.e. B is impossible) while $Bel_g(A \mid B)$ is undefined as soon as $Bel(B) = 0$ (i.e. B is unknown). This unability to update with a vacuous prior is very counterintuitive, with the geometric rule.

Another approach to conditioning has been proposed by De Campos et al. (1989) and Fagin & Halpern (1989) under the form

$$P^*(A \mid B) = \frac{Pl(A \cap B)}{Pl(A \cap B) + Bel(\overline{A} \cap B)} \quad (9)$$

$$P_*(A \mid B) = \frac{Bel(A \cap B)}{Bel(A \cap B) + Pl(\overline{A} \cap B)} \quad (10)$$

These definitions are justified by interpreting belief and plausibility functions as lower and upper probabilities, since it has been proved that

$$P^*(A \mid B) = \sup\{P(A \mid B) \mid P \in \mathcal{P}(Bel)\} \quad (11)$$

$$P_*(A \mid B) = \inf\{P(A \mid B) \mid P \in \mathcal{P}(Bel)\} \quad (12)$$

where $\mathcal{P}(Bel) = \{P \mid Bel(A) \leq P(A) \leq Pl(A), \forall A\}$. These conditional functions are actually upper and lower conditional probabilities and have been considered by Dempster himself. Although very satisfying from a probabilistic point of view, these definitions lead to a rather uninformative conditioning process since $P^*(\cdot \mid B) \geq Pl(\cdot \mid B) \geq Bel(\cdot \mid B) \geq P_*(\cdot \mid B)$. Especially, complete ignorance is obtained ($P^*(A \mid B) = 1$, $P_*(A \mid B) = 0$) as soon as $Bel(\overline{A} \cap B) = 0$ and $Bel(A \cap B) = 0$, i.e. as soon as the conditioning set B refines the granularity of the prior evidence by producing smaller focal elements. In that case the updating process corresponds to oblivion rather than learning.

Although difficult to justify from the point of view of upper and lower probability, Dempster rule of conditioning is more informative (increasing the precision of focal elements is permitted). Moreover this rule can be viewed as the intersection of the random set underlying Bel, and the conditioning set, i.e. it is completely justified from the standpoint of random sets and corresponds to a conjunctive set-theoretic operation. Then normalization is justified if the conditioning set must be taken for granted. Note that from the point of view of belief functions, the upper-lower probability view makes no sense just because belief functions are supposed to reflect a degree of certainty that uses a convention differing from probability functions ($Bel(A) = 1$ means certainty, $Bel(A) = 0$ means uncertainty) and that is *not* viewed as a lower probability (although from a mathematical point of view it is so). This point, i.e. that any set function can be used to represent certainty (up to further foundational issues) *without* referring to an unreachable probability function has often been overlooked by belief function opponents. Belief functions can be used as a model for evaluating certainty (this view is advocated by Smets (1988)) or as a model for capturing imprecision in probability (this view is that of Fagin & Halpern (1989), among others). Adopting the first point of view, Dempster rule of conditioning can be justified from a set of intuitive axioms (e.g. Cox conditioning axioms (Dubois & Prade, 1988b) or, the approach by Smets (1988)) that never uses the set of probabilities underlying the mathematical model of the belief functions.

Dempster rule of combination can be defined as a normalized intersection of two independent random sets $(\mathcal{F}_1, m_1)$ and $(\mathcal{F}_2, m_2)$

$$m(B) = [m_1 \oplus m_2](B) = \frac{\sum_{A_1 \cap A_2 = B} m_1(A_1) \cdot m_2(A_2)}{\sum_{A_1 \cap A_2 \neq \emptyset} m_1(A_1) \cdot m_2(A_2)} \quad (13)$$

This rule has been justified by Smets (1988, 1990) from axiomatic arguments. When the random set $(\mathcal{F}_2, m_2)$ associated with $m_2$ reduces to the ordinary subset A, i.e. $m_2(A) = 1$ and $\forall A' \neq A$, $m_2(A') = 0$, it can be easily checked that (13) and (4) give (6). Thus (13) extends (6) to the case of an uncertain observation represented by $(\mathcal{F}_2, m_2)$, but in a symmetrical manner. This is unfortunate from an updating point of view. Indeed Dempster rule embodies the combination of information from parallel sources that play the same role, while the notion of updating is basically dissymetrical: new information does not play the same role as a priori information. A non-symmetrical extension of (6) in case of uncertain observation, in the spirit of Jeffrey's rule (3), is provided by the formula

$$Pl(B \mid (\mathcal{F}_2, m_2)) = \sum_{A \subseteq \Omega} m_2(A) \cdot Pl_1(B \mid A) \quad (14)$$

where $Pl_1(B \mid A) = \frac{Pl_1(A \cap B)}{Pl_1(A)}$. The expression (14) can be interpreted in the following way: the subset A is the accurate description of what is observed with probability $m_2(A)$ and (14) is nothing but the expected plausibility of B given the uncertain observation. Formula (14) was suggested by Dubois & Prade (1986b; p. 140) up to a normalization factor and further discussed by Ichihashi & Tanaka (1989) among different alternatives to Dempster's rule. (13) and (14) coincide when the normalization factor of Dempster rule is 1.

The counterparts of (14) in terms of functions m or Bel can be easily obtained since it can be checked that the convex combination $\sum_{i=1,n} \alpha_i \cdot Pl_i$ with $\sum_{i=1,n} \alpha_i = 1$ corresponds to the plausibility function generated by the basic probability assignment $\sum_{i=1,n} \alpha_i \cdot m_i$ and is the dual, in the sense of (5), of the belief function defined by $\sum_{i=1,n} \alpha_i \cdot Bel_i$ (where $Bel_i$, as well as $Pl_i$ is defined from $m_i$). Thus we have

$$m(B \mid (\mathcal{F}_2, m_2)) = \sum_{A \subseteq \Omega} m_2(A) \cdot m_1(B \mid A) \quad (15)$$

where

$$m_1(B \mid A) = \left(\frac{1}{Pl_1(A)}\right) \cdot \sum_{\emptyset \neq B = C \cap A} m_1(C) \quad (16)$$

is the normalized basic probability assignment associated with $Pl_1(\cdot \mid A)$ (indeed $\sum_{B \subseteq \Omega} m_1(B \mid A) =$

1). In terms of belief functions we have
$$Bel(B \mid (\mathcal{F}_2, m_2)) = \sum_{A \subseteq \Omega} m_2(A) \cdot Bel_1(B \mid A) \quad (17)$$
where
$$Bel_1(B \mid A) = \frac{Bel_1(B \cup \overline{A}) - Bel_1(\overline{A})}{1 - Bel_1(\overline{A})} \quad (18)$$

It can be easily checked that (14) or (17) reduce to Jeffrey's rule when $(\mathcal{F}_1, m_1)$ defines a probability measure (i.e. $\mathcal{F}_1$ only contains singletons of $\Omega$) and $(\mathcal{F}_2, m_2)$ is such that $\mathcal{F}_2$ is a partition of $\Omega$. Wagner (1989) has recently established that the only eventwise combination of plausibility (or belief functions) is the linear convex combination, as it is the case for probability measures. This is a formal justification for (14) or (17), since as soon as we have in mind the random set view of a basic probability assignment, it is natural to require that the (plausibility or belief) function conditionalized by a random event depends only on the different conditional functions induced by the different realizations of this event.

It is important to point out that conditioning is meaningful only when observation does not completely contradict a priori knowledge. This is the case for Bayes rule where $P(B \mid A)$ is defined only if $P(A) > 0$ or for Dempster rule of conditioning where $Pl(B \mid A)$ is defined only if $Pl(A) > 0$. This is still the case for Jeffrey's rule where in (3) we should have $P(A_i) > 0$ as soon as $\alpha_i > 0$, as well as for its extended version (8) which is defined only if $\forall A$, $m_2(A) > 0$, $\exists C$, $m_1(C) > 0$ and $A \cap C \neq \emptyset$ (i.e. $Pl_1(A) > 0$). Note that Dempster's rule of combination is less requiring since it is still defined when $\exists A$, $m_2(A) > 0$ and $Pl_1(A) = 0$ (provided that it is not true for *all* A); it may seem a bit disturbing since it allows that the new information states, as somewhat probable, something which was held as certainly false according to previous information. We now examine the difference in behaviour of Dempster's rule of combination and of the extended Jeffrey's rule on a small example.

<u>Example</u> : Let $\mathcal{F}_1 = \{A_1, B_1\}$ with $m_1(A_1) = \alpha$ and $m_1(B_1) = 1 - \alpha$, $\mathcal{F}_2 = \{A_2, B_2\}$ with $m_2(A_2) = \beta$ and $m_2(B_2) = 1 - \beta$. Let us assume that $A_1 \cap A_2 = \emptyset$ ; $A_1 \cap B_1 \neq \emptyset$ ; $A_1 \cap B_2 \neq \emptyset$ ; $B_1 \cap B_2 \neq \emptyset$ ; $A_2 \cap B_1 \neq \emptyset$ ; $A_2 \cap B_2 \neq \emptyset$.

Dempster's rule yields $m = m_1 \oplus m_2$ with
$$m(A_1 \cap B_2) = \frac{\alpha \cdot (1 - \beta)}{1 - \alpha \cdot \beta} ; m(B_1 \cap A_2) = \frac{(1-\alpha) \cdot \beta}{1 - \alpha \cdot \beta} ; m(B_1 \cap B_2) = \frac{(1-\alpha)(1-\beta)}{1 - \alpha \cdot \beta}$$

while the extended Jeffrey's rule gives
$$m(B_1 \cap A_2 \mid A_2) = \frac{1 - \alpha}{1 - \alpha} = 1 \text{ if } \alpha \neq 1 ;$$
$$m(A_1 \cap B_2 \mid B_2) = \alpha ; m(B_1 \cap B_2 \mid B_2) = 1 - \alpha$$
and finally
$$m(B_1 \cap A_2 \mid (\mathcal{F}_2, m_2)) = \beta ; m(A_1 \cap B_2 \mid (\mathcal{F}_2, m_2)) = \alpha(1 - \beta) ; m(B_1 \cap B_2 \mid (\mathcal{F}_2, m_2)) = (1 - \alpha)(1 - \beta).$$

As it can be seen on this example, and easily proved in the general case from (15)-(16), the basic probability assignments obtained by Dempster's rule and the extended Jeffrey's rule have exactly the same focal elements but their weights are different. Moreover the extended Jeffrey's rule gives a non-symmetrical result as expected. When $\alpha = 1$ this latter rule does not apply since then $\mathcal{F}_1 = \{A_1\}$ and one of the focal elements of $\mathcal{F}_2$, namely, $A_2$ is such that $Pl_1(A_2) = 0$ due to $A_1 \cap A_2 = \emptyset$. In this case Dempster's rule gives $m(A_1 \cap B_2) = 1$ whatever the value of $\beta$ (provided that $\beta \neq 1$), i.e. a conclusion which is not pervaded with uncertainty in spite of the fact we may have a strong conflict between $\mathcal{F}_1$ and $\mathcal{F}_2$ if $\beta$ is close to 1 (then $1 - \alpha\beta$ is close to 0). Also in the example, the extended Jeffrey's rule looks more robust partly because it does not apply when the behaviour of Dempster's rule is particularly questionable, and also because the normalization is performed in a global way in Dempster's rule, while in the other case it takes place at the level of each focal elements of the body of evidence corresponding to the uncertain observation upon which we conditionalize.

Figure 1 provides a summary of the relationships between the various rules. It is clear that we can define a Jeffrey-like extension of the geometric rule mentioned above (by substituting in (17) the alternative definition of $Bel_1(B \mid A)$), and more generally, definitions like (14) or (17) are compatible with the use of other definitions of conditioning in case of a sure observation.

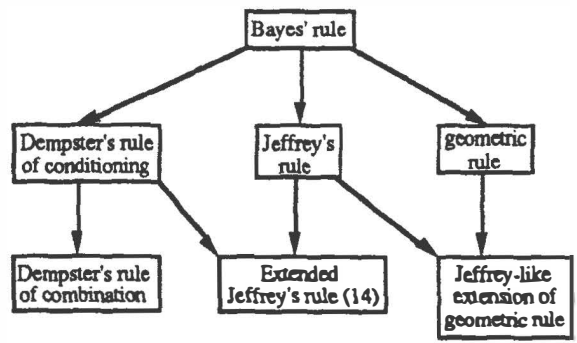

Figure 1 : Generalizations of Bayes' rule

## 3 - Updating in possibility theory

Possibility measures and necessity measures are respectively particular cases of plausibility functions and belief functions when the focal elements are nested. However the linear convex combination of possibility measures (or of necessity measures) does not yield a possibility measure (or a necessity measure) generally (see Dubois & Prade, 1986a). Thus the approach presented in the preceding section cannot be applied to possibility and necessity measures for updating them under an uncertain observation (expressed in a possibilistic way), if we further require that the result be a possibility measure. Let us first recall that a possibility measure $\Pi$ over $\Omega$ can be defined, through a so-called possibility distribution $\pi$, which is a function from $\Omega$ to $[0,1]$, by the formula (Zadeh, 1978)

$$\forall A \subseteq \Omega, \Pi(A) = \sup_{\omega \in A} \pi(\omega) \quad (19)$$

where $\pi(\omega)$ estimates to what extent it is possible (i.e. compatible with what is known) that $\omega$ corresponds to the true state of the reality in $\Omega$. In other words $\pi$ restricts the more or less possible states of the reality given the available incomplete information about the reality $\pi$ is supposed to be normalized, i.e. $\exists \omega \in \Omega, \pi(\omega) = 1$ in other words, $\Omega$ is supposed to be an exhaustive set of alternatives, one of which is completely posible. To $\Pi$ is associated a necessity measure $N$, by duality, namely

$$\forall A \subseteq \Omega, N(A) = 1 - \Pi(\overline{A}) = \inf_{\omega \notin A} [1 - \pi(\omega)] \quad (20)$$

Recently, it has been shown (Dubois & Prade, 1990) that the only way of combining possibility measures $\Pi_1, ..., \Pi_n$ into a possibility measure $\Pi$, in an eventwise manner, was a max-combination of the form

$$\forall A, \Pi(A) = \max(f_1(\Pi_1(A)), ..., f_n(\Pi_n(A))) \quad (21)$$

where $f_i$ is a monotonically increasing function such that $f_i(0) = 0, \forall i$ and $\exists j, f_j(1) = 1$ which modifies the shape of the possibility distribution $\pi_i$ underlying $\Pi_i$. An example of admissible possibility consensus function is the weighted maximum operation, i.e.

$$\Pi(A) = \max_{j=1,n} \min(\lambda_j, \Pi_j(A)) \quad (22)$$

with $\max_{j=1,n} \lambda_j = 1$, where $\lambda_j$ represents the relative importance of the source yielding $\Pi_j$ (Dubois & Prade, 1986c). However, in (22), the minimum can be changed into a product, or into the linear operation $\max(0, a + b - 1)$, and more generally into any operation $*$ with $1 * 1 = 1$, $0 * 1 = 0 = 1 * 0$, and increasing in both places.

In fact, the weighted max-combination is the counterpart in possibility theory of the linear convex combination in probability theory; the weighted max-combination can be interpreted in the possibilistic framework just as the convex combination can be interpreted in terms of probabilistic expectation. This leads to the following updating formula for an a priori possibility distribution $\pi_1$ in the face of a new piece of information in the form of another possibility measure $\pi_2$. It can be expressed in terms of possibility distributions

$$[\pi_1 | \pi_2](\omega) = \sup_{\alpha \in (0,1]} \min(\alpha, \pi_1(\omega | B_{2\alpha})) \quad (23)$$

and in terms of possibility measures

$$[\Pi_1 | \Pi_2](A) = \sup_{\alpha \in (0,1]} \min(\alpha, \Pi_1(A | B_{2\alpha})) \quad (24)$$

The observation (pervaded with uncertainty) represented by $\pi_2$ is here viewed as a weighted family of observations in the usual sense, i.e. the weight $\alpha$ reflects to what extent $B_{2\alpha}$ is an admissible crisp representation of the fuzzy set $B_2$ (such that $\mu_{B_2} = \pi_2$). Namely we have (Zadeh, 1971)

$$\mu_{B_2}(\omega) = \sup_{\alpha \in (0,1]} \min(\alpha, \mu_{B_{2\alpha}}(\omega)) \quad (25)$$

where $B_{2\alpha}$ is the $\alpha$-level-cut of $B_2$, namely $B_{2\alpha} = \{\omega \in \Omega, \mu_{B_{2\alpha}}(\omega) \geq \alpha\}$. A "sup" in (23)-(24) is used since, in the general case, there an infinite number of distinct level-cuts $B_{2\alpha}$. The counterpart of formula (24) in terms of necessity measure is

$$[N_1 | \Pi_2](A) =$$
$$\inf_{\alpha \in (0,1]} \max(1 - \alpha, N_1(A | B_{2\alpha})) \quad (26)$$

In other words $\{(B_{2\alpha}, \alpha) | \alpha \in (0,1]\}$ can be viewed as a basic *possibilistic* assignment, and $\alpha$ is indeed the possibility that the "possibilistic set" attached to $\pi_2$, (just as a random set is attached to a basic probability assignment) is precisely equal to $B_{2\alpha}$. It is worth noticing that the expressions defining $[\pi_1 | \pi_2]$ or $[\Pi_1 | \Pi_2]$ are integrals in the sense of (Sugeno, 1977) when $\omega$ or $A$ is fixed, just as Jeffrey's rule viewed as an expectation, is an integral in the usual sense (in a finite setting). Observe that the rule of conditioning gives the maximal importance to the core of the fuzzy set $B_2$ (the set of elements with membership 1) which, being the smallest level cut, is the more informative, and less and less importance when level cuts become larger ($\alpha < \beta \Rightarrow B_{2\alpha} \supseteq B_{2\beta}$).

We have now to give the Bayes-like rule of conditioning in possibility theory, i.e. a possibility distribution $\pi_1$ conditionalized by $B$ results in the possibility distribution $\pi(\cdot | B)$, defined in accordance with Dempster rule of conditioning, by

$$\pi(\omega | B) = \frac{\pi_1(\omega)}{\Pi_1(B)} \text{ if } \omega \in B \quad (27)$$
$$= 0 \text{ otherwise}$$

where $\Pi_1(B) = \sup_{\omega \in B} \pi_1(\omega)$ and

312

$$\Pi(A|B) = \frac{1}{\Pi_1(B)} \sup_{\omega \in A} \pi_1(\omega) = \frac{\Pi_1(A \cap B)}{\Pi_1(B)} \quad (28)$$

together with $N(A | B) = 1 - \Pi(\bar{A} | B)$. See (Dubois & Prade, 1988b) for a discussion of conditioning in possibility theory and a justification of this formula. Note also that the rule of conditioning in possibility theory is a particular case of the more general symmetric rule of combination (see Dubois & Prade (1988a), Shafer (1987) for instance)

$$\pi(\omega) = \frac{\pi_1(\omega) * \pi_2(\omega)}{\sup_{\omega \in \Omega} (\pi_1(\omega) * \pi_2(\omega))} \quad (29)$$

where * is a conjunctive operation which is symmetrical, non-decreasing, and such that $\forall a \in [0,1]$, $a * 1 = a$. This rule is the possibilistic counterpart of Dempster rule of combination, while (24) plays the same role with respect to Jeffrey's updating rule.

Introducing (27) into (24) leads to the following updating formula
$[\pi_1 | \pi_2](\omega) =$
$$\sup_{\alpha \in (0,1]} \min(\alpha, \frac{\pi_1(\omega)}{\Pi_1(B_{2\alpha})}, \mu_{B_{2\alpha}}(\omega)) \quad (30)$$

Observe that when $\alpha$ decreases, $\Pi_1(B_{2\alpha})$ can only increase (since $B_{2\alpha}$ become larger) and thus $\frac{\pi_1(\omega)}{\Pi_1(B_{2\alpha})}$ can only decrease. Moreover, $\mu_{B_{2\alpha}}(\omega) = 1$ only if $\alpha \leq \pi_2(\omega)$ and 0 otherwise. Hence the suprenum in (30) is attained for $\alpha = \pi_2(\omega)$. The updating formula can thus be expressed in a more compact way:

$$[\pi_1 | \pi_2](\omega) = \min(\pi_2(\omega), \frac{\pi_1(\omega)}{\Pi_1(B_{2\pi_2(\omega)})}) \quad (31)$$

where $B_{2\pi_2(\omega)} = \{\omega' | \pi_2(\omega') \geq \pi_2(\omega)\}$. The effect of this updating formula is pictured on Figure 2.

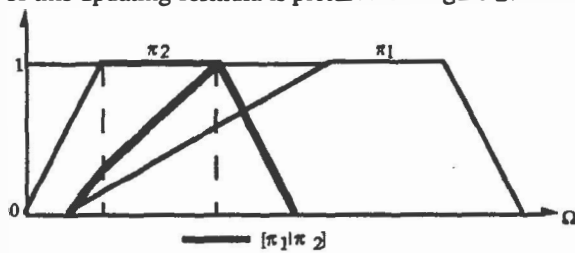

Figure 2

It is worth noticing that $[\pi_1 | \pi_2]$ is normalized as soon as the core of $\pi_2$ overlaps the support of $\pi_2$ (i.e. $\exists \omega \in \Omega, \pi_2(\omega) = 1$ and $\pi_1(\omega) > 0$). This is satisfying; if it is not the case, it would mean that the main part of $B_2$ focuses on values which are completely impossible according to $\pi_1$ and then the conditionalization is debatable. The conditionalization becomes completely undefined in case of total conflict between $\pi_2$ and $\pi_1$ (i.e. when it does not exist $\omega$ such that $\pi_1(\omega) > 0$ and $\pi_2(\omega) > 0$).

Another important property is that $[\pi_1 | \pi_2] = \min(\pi_1, \pi_2)$ when the cores of $\pi_1$ and $\pi_2$ are overlapping. This is similar to the coincidence between the extended Jeffrey's rule and Dempster rule when no normalization factor is necessary in the latter. This is well in accordance with the fact that if the available information is of the form $x \in A$, then upon arrival of a sure piece of information $x \in B$, the updating process consists in producing $x \in A \cap B$. More generally the denominator $\Pi_1(B_{2\pi_2(\omega)})$ in (31) helps producing a normalized result on the basis that $\pi_2$ is considered as certain, in the spirit of conditioning.

Let us examine the particular case where
$$\pi_2(\omega) = \max(\mu_B(\omega), \lambda)$$
where B is an ordinary subset of $\Omega$. Then it means that B is completely possible ($\Pi_2(B) = 1$) and there is a possibility equal to $\lambda$ that the observation is outside B ($\Pi_2(\bar{B}) = \lambda$). In that case the result of the conditionalization of $\pi_1$ by the uncertain observation represented by $\pi_2$ is given by

$$[\pi_1|\pi_2](\omega) = \max(\min(\frac{\pi_1(\omega)}{\Pi_1(B)}, \mu_B(\omega)), \min(\lambda, \pi_1(\omega)))$$

This is illustrated by Figure 3, where we see that in that case the result is the intersection of $\pi_1$ and $\pi_2$, which is renormalized over the subset B only. As expected, $[\Pi_1|\Pi_2](\bar{B}) = \lambda$, and the combination is dissymetrical; it favors the new information ($\pi_2$) over the old one.

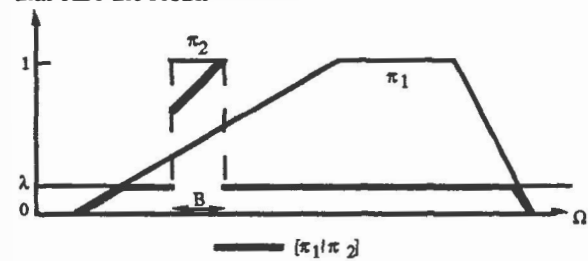

Figure 3

### 4 - Possibility theory and Spohn's ordinal conditional functions

An ordinal conditional function (OCF for short) is a function $\kappa$ from a complete field of propositions into the class of ordinals. Here, for simplicity we consider a function from a finite Boolean algebra B to the set of natural integers n. This Boolean algebra consists of a family of subsets of a universe $\Omega$ induced by a finite partition $\{A_1, ..., A_m\}$ of $\Omega$. By definition an OCF verifies the following properties

(Spohn, 1988a,b) :
i) $\forall i, \forall \omega, \omega' \in A_i, \kappa(\omega) = \kappa(\omega')$
ii) $\exists A_i \subseteq \Omega, \kappa(A_i) = 0$
iii) $\forall A \subseteq \Omega, \kappa(A) = \min\{\kappa(\omega) \mid \omega \in A\}$.

It is easy to see that the set function $N\kappa$ defined by $N_\kappa(A) = 1 - e^{-\kappa(\bar{A})}$ is a necessity measure, with values in a subset of the unit interval. Moreover because $\kappa(A) \in \mathbb{N}$, $N_\kappa(A) < 1$, $\forall A \neq \Omega$. The set $\{\kappa(\omega) \mid \omega \in \Omega\}$ is the counterpart of a possibility distribution $\pi$ on $\Omega$, such that $\Pi(A) = \max\{\pi(\omega) \mid \omega \in A\}$. Namely, let $\Pi_\kappa(A) = 1 - N_\kappa(\bar{A}) = e^{-\kappa(A)}$, it is easy to check that $\pi_\kappa(\omega)$ is equal to $e^{-\kappa(\omega)}$, where $\pi_\kappa$ is the possibility distribution associated with $\Pi_\kappa$. $\kappa(\omega)$ can be viewed as a degree of impossibility of $\omega$, and $\kappa(A) = 0$ means A is completely possible. Since $\kappa(\omega) \in \mathbb{N}$, $\pi_\kappa(\omega) > 0$ for all $\omega$'s, i.e. nothing is considered as fully impossible in Spohn's approach.

Spohn (1988) also introduces conditioning concepts, especially :
- the A-part of $\kappa$ such that
$$\forall \omega \in A, \kappa(\omega \mid A) = \kappa(\omega) - \kappa(A) \quad (32)$$
- the $(A,n)$-conditionalization of $\kappa$, say $\kappa(\omega \mid (A,n))$ defined by
$$\kappa(\omega \mid (A,n)) = \kappa(\omega \mid A) \text{ if } \omega \in A$$
$$= n + \kappa(\omega \mid \bar{A}) \text{ if } \omega \in \bar{A} \quad (33)$$

It is interesting to translate this notion into the possibilistic setting. Definitions (32) and (33) respectively become

$$\pi_\kappa(\omega \mid A) = \frac{\pi_\kappa(\omega)}{\Pi_\kappa(A)} \text{ if } \omega \in A \quad (34)$$

$$\pi_\kappa(\omega \mid (A,n)) = \frac{\pi_\kappa(\omega)}{\Pi_\kappa(A)} \text{ if } \omega \in A \quad (35)$$

$$= e^{-n} \cdot \frac{\pi_\kappa(\omega)}{\Pi_\kappa(\bar{A})} \text{ if } \omega \notin A$$

It is trivial to check that (34) corresponds to Dempster rule of conditioning but (35) does not correspond to Dempster rule of combination although (34) is a particular case of (35) for $n \to +\infty$ (it says then that $\bar{A}$ is considered as impossible). In fact (34) is exactly the Bayes-like conditioning of possibility measures, i.e. equation (27).

In order to compare Spohn's rule to the possibilistic updating rule, we let $\alpha = e^{-n}$ and note that it comes down to updating a possibility distribution $\pi_1 = \pi_\kappa$ on the basis of an uncertain observation $\Pi(\bar{A}) = \alpha$ (or equivalently $N(A) = 1 - \alpha$ in terms of the degree of certainty). A comparison of the two rules is given in Figure 4 for the three mutually exclusive and exhaustive cases $A \cap C = \emptyset$, $C \subset A$ and C overlaps A but is not contained in A, where C is the core of $\pi_1$.

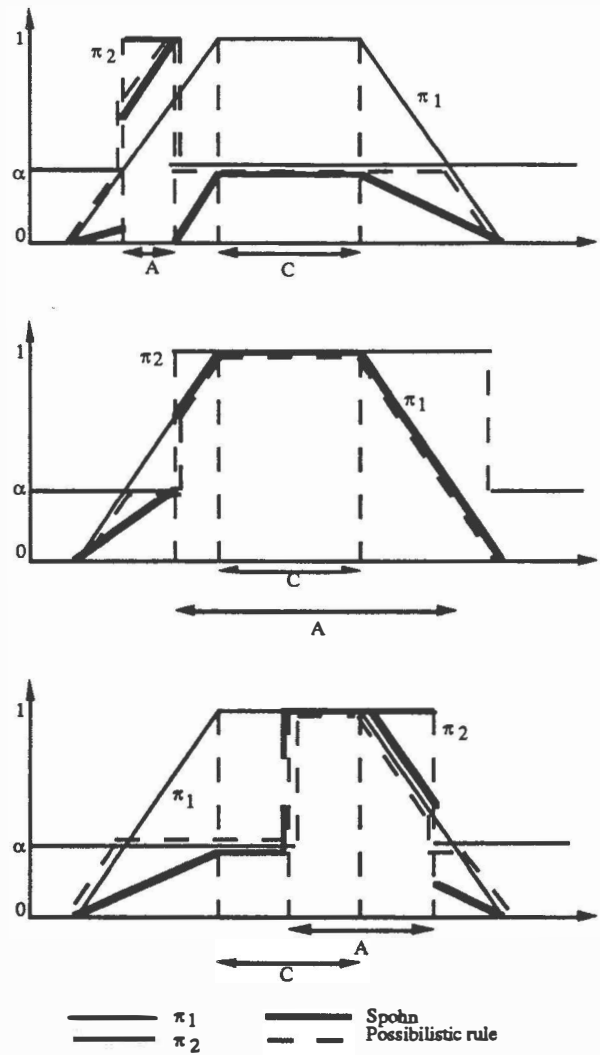

——— $\pi_1$ ——— Spohn
——— $\pi_2$ - - - Possibilistic rule

Figure 4

Some comments are in order. First, the two rules apparently produce very similar results, except that Spohn's rule retains the shape of $\pi_1$ on $\bar{A}$ more faithfully than the possibilistic rule. The difference can be attenuated on the example, by substituting a product rule to the minimum rule in (31). However the comparison is limited by the fact that the possibilistic rule applies in more cases than Spohn's rule does. In order to make a more extended comparison, Spohn's rule must be generalized to the updating of a possibility distribution $\pi_1$ by another distribution $\pi_2$. (34) and (35) can be straightforwardly extended to a partition $\{A_1, A_2, ..., A_n\}$ of $\Omega$, with $\alpha_i = \Pi_2(A_i)$ as follows :

$$\pi(\omega \mid \{(A_i, \alpha_i) \mid i=1,n\}) = \frac{\alpha_i \cdot \pi_1(\omega)}{\Pi_1(A_i)} \text{ for } \omega \in A_i, i=1,n \quad (36)$$

with the condition $\max_i \alpha_i = 1$ (normalization of $\Pi_2$). (34) and (35) are obtained letting $A_1 = A$, $\alpha_1 = 1$, $A_2 = \bar{A}$, $\alpha_2 = \alpha$. A special case of (36) is when we choose the set of singletons of $\Omega$ as the partition, and $\pi_2(\omega)$ instead of the $\alpha_i$'s. Then (36) obviously gives

$$\pi(\omega \mid \pi_2) = \pi_2(\omega), \forall \omega \in \Omega$$

i.e. Spohn's rule comes down to changing $\pi_1$ into $\pi_2$ in a systematic way, for all $\omega$ such that $\pi_1(\omega) > 0$ (and Spohn's assumptions are such that this is the case for all $\omega \in \Omega$). This is exactly the same as for Jeffrey's rule in the probabilistic setting.

The difference between the two rules becomes patent. Spohn's rule always strongly accounts for the new information, possibly forgetting the old one, even if it was more precise (i.e. $\pi_2 \geq \pi_1$). On the contrary the possibilistic rule always tries to precisiate the new information by means of the old one when it applies. Particularly we have the following noticeable property : if $\pi_2 \geq \pi_1$ then the possibilistic rule gives $[\pi_1 \mid \pi_2] = \pi_1$ (while Spohn's rule prefers $\pi_2$). Indeed if $\pi_2 \geq \pi_1$ then $\Pi_1(B_{2\pi_2(\omega)})=1$, $\forall \omega$, in (31). Particularly if in the above example where $\Pi_2(\bar{A}) = \alpha$, if $\pi_1 \leq \max(\mu_A, \alpha)$ then the possibilistic rule would leave $\pi_1$ unchanged because $\Pi_1(\bar{A}) < \alpha$, i.e. the input information is too weak to question the available knowledge. This behavior, i.e. productive updating when the new information is not already entailed by the old one is very natural since, when $\pi_2 \geq \pi_1$, it means that $\pi_2$ tells the same as $\pi_1$ but is only weaker. This is exactly what happens when in a propositional knowledge base a redundant proposition is added. Spohn's rule (extended by (36)) violates this requirement ; it corresponds to the idea that the new information must be kept even if it leads us to forget something that is already known. Conversely, if $\pi_2 \leq \pi_1$, then both rules produce $\pi_2$ as the updated knowledge, because $\pi_2$ is completely consistent with $\pi_1$ but more precise.

## Conclusion

In this paper, we have demonstrated that there are two kinds of rules for the combination of information : symmetric rules (that combine sources in parallel) and dissymmetric rules that correspond to the idea of updating. Dempster rule and fuzzy set intersections are among the first kind of rules while Jeffrey's rule, is of the other kind, as well as the rules proposed in this paper for belief functions and possibility measures. These two kinds of uncertainty measures lead to different updating formulas only for the sake of preserving closure conditions with respect to the theories. The difference between the possibilistic rule and Spohn's rule has been studied, using a scale transformation that changes an ordinal conditional function into a possibility measure. This examination has pointed out that there may be two kinds of updating rules : the ones that preserve the available knowledge while possibly refining it, the ones that systematically consider the a priori knowledge as being kind of obsolete in the face of new information. The possibilistic rule and Spohn's rule correspond to the first and the second point of view respectively. Interestingly Shenoy (1989) has recently discussed Spohn's rule in comparison with Dempster rule of combination ; our paper and his are thus complementary in this respect.

Lastly the introduced rules have been constructed on the basis of analogy with Jeffrey's rule. But the latter can be justified in terms of principles of minimum change ; namely the result of Jeffrey's rule is the probability distribution obtained by minimizing the relative entropy with respect to the prior probability, using the uncertain observations as constraints (Domotor, 1985). A further topic of research would be to look at information theoretic justifications of rules introduced in this paper, following the methodology outlined in a previous paper (see Dubois & Prade, 1987), for instance using information closeness indices recently introduced by Ramer (1989) in the possibilistic setting.

## Session 7:

## Dempster-Shafer: Graph Decomposition, FMT, Interpretations